\documentclass[10pt,journal,compsoc]{IEEEtran}
\pdfoutput=1

\usepackage{amsmath,amssymb,amsfonts}
\usepackage{float}
\usepackage{multirow}
\usepackage{epsfig}
\usepackage{graphicx}
\usepackage{algorithm}
\usepackage{algorithmic}
\usepackage{subfigure}
\usepackage{epstopdf}
\usepackage{makecell}
\usepackage{epstopdf}
\usepackage{bm}
\usepackage{times}
\usepackage{booktabs}
\usepackage{microtype}
\usepackage{cite}
\usepackage{color}

\usepackage{hyperref}
\usepackage{url}
\usepackage[justification=raggedright]{caption}

\ifCLASSINFOpdf
\else
\fi

\begin{document}
	
	\title{AGFormer: Efficient Graph Representation with Anchor-Graph Transformer}
	
	\author{{Bo Jiang, Fei Xu, Ziyan Zhang, Jin Tang and Feiping Nie}\IEEEcompsocitemizethanks{\IEEEcompsocthanksitem Bo Jiang, Fei Xu, Ziyan Zhang, Jin Tang are with the Anhui Provincial Key Laboratory of Multimodal Cognitive Computation, School of Computer Science and Technology of Anhui University, Hefei, 230601, China.\protect\\
			Feiping Nie is with the School of Artificial Intelligence, Optics and Electronics (iOPEN), and the Key Laboratory of Intelligent Interaction and Applications (Ministry of Industry and Information Technology), Northwestern Polytechnical University.\protect\\
			Corresponding author: Bo Jiang, E-mail: jiangbo@ahu.edu.cn}}

	\markboth{Journal of \LaTeX\ Class Files,~Vol.~14, No.~8, August~2015}%
	{Shell \MakeLowercase{\textit{et al.}}: Bare Demo of IEEEtran.cls for Computer Society Journals}
	
	\IEEEtitleabstractindextext{%
		\begin{abstract}
			
			To alleviate the local receptive issue of GCN, Transformers have been exploited  to capture the long range dependences of nodes for graph data representation and learning. 
			However, 
			existing graph Transformers generally employ regular
			self-attention module for all node-to-node message passing which needs to learn the
			affinities/relationships between all node’s pairs, leading to  high
			computational cost issue. Also, they  are usually sensitive to graph noises. 
			To overcome this issue, 
			we propose a novel
			graph Transformer architecture, termed Anchor Graph Transformer
			(AGFormer), by leveraging an anchor graph model. 
			To be specific, AGFormer first obtains some representative
			anchors and then converts node-to-node message passing into
			anchor-to-anchor and anchor-to-node message passing process. Thus, AGFormer performs much more efficiently and also robustly than regular node-to-node Transformers. 
			Extensive experiments on several benchmark
			datasets demonstrate the effectiveness and benefits of proposed AGFormer. 
			
		\end{abstract}
		
		\begin{IEEEkeywords}
			Graph convolutional network, Graph Transformer, Anchor graph, Graph representation learning
	\end{IEEEkeywords}}

	\maketitle
	\section{Introduction}
	
	Graph representation and learning 
	is an important problem in machine learning and data mining fields. 
	The goal of graph learning is to learn effective node representations for the downstream tasks, such as semi-supervised learning, graph classification and clustering etc. 
	Graph convolutional networks (GCNs)~\cite{spectralGCN,GCN,WuSurvey,TNNLS-GCoNN} have been demonstrated to be powerful on addressing graph data representation and learning tasks.
	For example, 
	Kipf et al.~\cite{GCN} propose Graph Convolutional Network (GCN) for graph data representation learning by exploiting 
	the spectral representation of graph. 
	Veli{\v{c}}kovi{\'{c}} et al.~\cite{GAT} propose Graph Attention Networks (GAT) which assigns the attention weights to the neighbors and then conducts the message aggregation on the attention-weighted graph.  
	Hamilton et al.~\cite{GraphSage} propose GraphSAGE which first samples some neighbors for each node and then  aggregates the information from them for contextual representation. 
	One can refer work~\cite{WuSurvey} for the more detailed survey. 
	However, as we all know that one main
	limitation of GCNs is that 
	they generally conduct
	message aggregation on local 
	neighbors which thus fail to capture the long range dependences of nodes. 
	Although deep multi-layer architecture can 
	enlarge the receptive field, however,  as we know that,  
	deep GCNs usually suffer from the over-smoothing issue~\cite{chen2020measuring}. 
	
	To overcome this limitation,  
	in recent years, Transformer models have been leveraged for graph representation and learning tasks. 
	The core of graph Transformer 
	is to utilize the self-attention mechanism to 
	capture the long-range depedences of nodes (tokens) for global contextual representation and learning. 
	For example, Wu et al.~\cite{Wu2021GraphTrans} propose GraphTrans which uses a regular 
	self-attention to capture the long-range relationships and employs a specific 'cls' token to obtain the global embedding for graph classification problem. 
	Zhang et al.~\cite{zhang2022hierarchical} propose Adaptive Node Sampling for Graph Transformer (ANS-GT) which designs some adaptive node sampling strategies to address the transformer's input length and capture the long-range dependences of nodes via self-attention. 
	Dwivedi et al.~\cite{dwivedi2020generalization} propose GraphTransformer (GT) which uses Laplacian eigenvectors to represent the location encoding and focuses on message passing in the self-attention module. 
	However, existing graph Transformers generally employ regular self-attention module for node-to-node message passing which needs to learn the  affinities/relationships between all node pairs. 
	This obviously leads to high computational cost  which limits its application on the large-scale graph  learning problem. 
	Also, they are usually sensitive to graph noises. 
	To overcome this issue, some recent works~\cite{gao2022patchgt,kuangcoarformer} suggest to 
	conduct 
	Transformer/self-attention learning on the coarse graph level, such as graph patches~\cite{gao2022patchgt}, communities~\cite{kuangcoarformer} etc. 
	However, the Transformers used in these approaches generally learn the coarse (or patch)-level representations, which fails to be fully aware of the original node
	representations in their learning process. 
	Therefore, how to employ Transformers for graph data representation and learning is still a challenge problem.  
	

	To address these issues, inspired by 
	Set Attention (ISA)~\cite{lee2019set}, in this paper, we propose a novel graph Transformer architecture, termed Anchor Graph Transformer (AGFormer), by leveraging an anchor graph model. 
	Anchor graph model has been studied in large-scale data mining problem, such as semi-supervised learning and clustering~\cite{nie2021fast}, image representation~\cite{chen2019deep}, to speed up the learning process. 
	Inspired by this, in this paper, we leverage it into graph Transformer architecture. To our best knowledge, anchor graph has not been studied or emphasized for graph Transformer representation. 
	The core idea of the proposed AGFormer is to first obtain some representative anchors 
	and then leverage these anchors as message bottleneck to learn the representations for all nodes. 
	To be specific, 
	AGFormer converts node-to-node message passing (in self-attention) into anchor-to-anchor and anchor-to-node message passing and therefore implements significantly more efficiently than regular node-to-node message passing, as illustrated in Figure 1. 
	Also, it is less sensitive to the outlier/noisy nodes. 
	Overall, 
	the proposed AGFormer mainly contains three modules, i.e., i) GCN  node embedding, ii)
	anchor-to-anchor  self-attention and iii) node-to-anchor  cross-attention. 
	We \textbf{first} adopt the multi-layer GCN module to learn the initial local neighbor-aware embeddings  for graph nodes. 
	\textbf{Then}, we design an anchor-to-anchor  self-attention mechanism to achieve message propagation across different anchors. 
	\textbf{Finally}, we adopt an anchor-to-node   cross-attention to conduct message propagation between anchors and nodes and obtain final node embeddings. 
	Comparing with existing graph Transformers, the proposed AGFormer  
	provides an efficient and robust way to learn node-level representations by integrating both local and global dependences  together. 
	
	Overall, we summarize the main contributions of this paper as follows,  
	
	\begin{itemize}
		\item We propose to leverage anchor graph model into Transformer architecture and develop a simple yet efficient AGFormer to achieve long-range learning on graph. 
		
		\item We propose to joint graph 
		convolution and AGFormer together to 
		present a new learning architecture  for graph data. 
		The proposed approach captures both local receptive field and long-range dependences of nodes simultaneously for graph data representation.  
		
		\item Experiments on four widely used benchmark datasets demonstrate the effectiveness, efficiency and robustness of our proposed AGFormer approach.  
	\end{itemize}
	
	\begin{figure}[htpb]
		\includegraphics[width=0.49\textwidth]{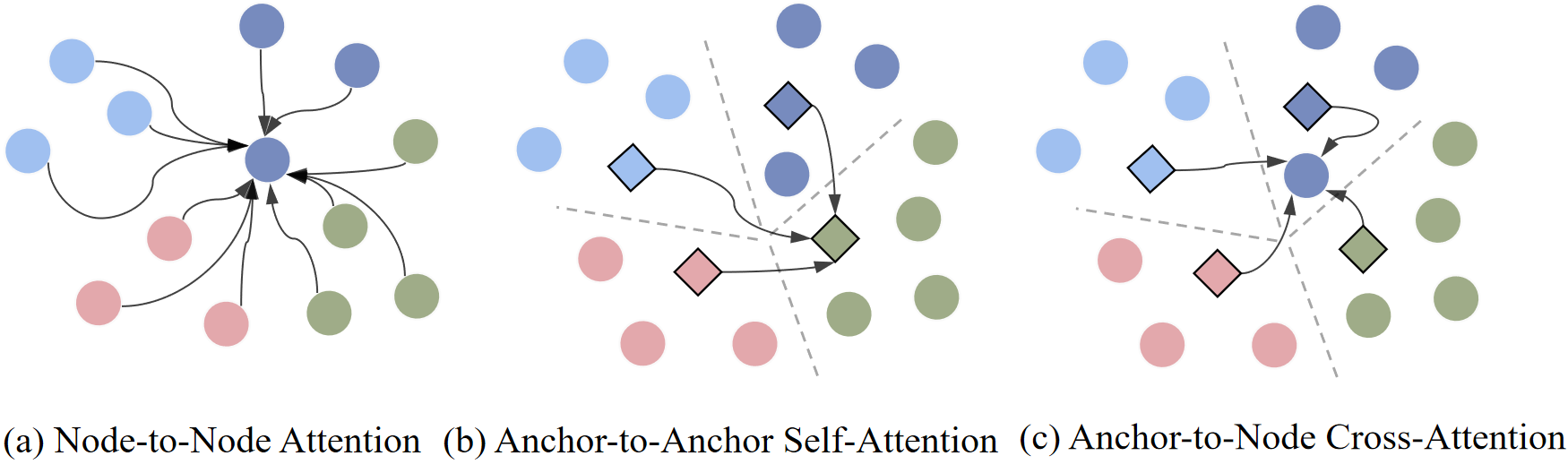}
		\caption{ Block diagram of AGFormer with information interaction.  (a) Node-to-node information interaction in regular Transformer.  (b) Anchor-to-anchor  interaction in AGFormer. (c) Anchor-to-node interaction in AGFormer.  
		}
		\label{fig:attention_module}
	\end{figure}

	\section{Related Works}
	\subsection{Graph Convolutional Network} 
	Graph Convolutional Networks have been successfully applied on various graph learning tasks, such as node classification~\cite{bhagat2011node,GraphSage,GCN,GAT,DropEdge}, graph classification~\cite{duvenaud2015convolutional,lee2019self,ying2018hierarchical,li2015gated}, 
	link prediction~\cite{kipf2016variational,schlichtkrull2018modeling,zhang2018link}, etc. 
	Scarselli et al.~\cite{GNN} propose Graph Neural Network (GNN) which processes both node and graph-level learning tasks simultaneously. 
	Kipf et al.~\cite{GCN} propose the widely used Graph Convolutional Network (GCN).
	Hamilton et al.~\cite{GraphSage} propose GraphSAGE which develops a neighborhood sampling strategy for processing large-scale graph data.
	Veli{\v{c}}kovi{\'{c}} et al.~\cite{GAT} propose Graph Attention Networks (GATs) which achieves the adaptive assignment of weights to different neighbors through the multi-head self-attention mechanism. 
	Wu et al.~\cite{SGC} propose more efficient Simple Graph Convolution (SGC) which converts the nonlinear GCN into a single linear transformation by  removing nonlinear activation layers.
	Yang et al.~\cite{FactorGCN} propose Factorizable Graph Convolutional network (FactorGCN), which disentangles the  simple graph into several subgraphs of potential relationships to produce disentangled features.
	Jiang et al.~\cite{GLCN} propose Graph Learning-Convolutional Network (GLCN). It combines graph learning and graph convolution together in a unified network structure to learn an optimal graph representation for semi-supervised learning task.
	Jin et al.~\cite{prognn} propose Property GNN (Pro-GNN) which learns clean graph structure to defend against adversarial attacks. 
	Yang et al.~\cite{yang2022ncgnn} propose Node-level Capsule Graph Neural Network (NCGNN). 
	Zhu et al.~\cite{gcnlhf} design a unified optimization objective framework GNN-LF/HF with adjustable convolution kernels representing both low-pass and high-pass filters. 
	Jiang et al.~\cite{GECN} propose Graph elastic Convolution Network (GeCN) which integrates elastic net selection into graph convolution for robust graph representation. 
	
	\subsection{Graph Transformers} 
	Due to its capability to represent the long-range relationships, many works consider applying Transformers for graph data representation and learning tasks.  
	For example, 
	Ying et al.~\cite{NEURIPS2021_Graphormer} propose Graphormer which encodes the edge information into Transformer to perceive the structure of the graph. 
	Rong et al.~\cite{rong2020GROVER} propose GROVER which aims to capture the rich semantic and structural information in molecules from a large amount of unlabeled data. 
	Wu et al.~\cite{Wu2021GraphTrans} propose Graph Transformer (GraphTrans) which uses node-to-node self-attention to learn long-range pairwise relationships.
	Nguyen et al.~\cite{nguyen2022UGformer} propose Universal Graph Transformers (UGformers) for robust graph representation. 
	Kreuzer et al.~\cite{kreuzer2021rethinking} propose a Spectral Attention Network (SAN), which rethinks the graph transformer with spectral attention and learns a position encoding for each node based on the eigenvalues and eigenvectors of the Laplacian matrix.
	Chen et al.~\cite{chen2022SAT} propose Structure-Aware Transformer (SAT),
	which simultaneously utilizes both node and subgraph tokens to capture the local structural information of graph.
	Ramp{\'a}{\v{s}}ek et al.~\cite{rampavsek2022recipe} propose a general framework, namely General, Powerful, and Scalable graph Transformer (GPS), which decouples the edge aggregation from the fully connected Transformer to reduce the complexity. 
	Kim et al.~\cite{kim2022pure} propose Tokenized Graph Transformer (TokenGT). It  treats all nodes and edges in the graph as independent tokens which are fed into the transformer. 
	Zhang et al.~\cite{zhang2022hierarchical} propose Adaptive Node Sampling for Graph Transformer (ANS-GT) which introduces a hierarchical attention scheme with graph coarsening to capture the long-range dependencies. 
	Some recent works~\cite{gao2022patchgt,kuangcoarformer} also suggest to 
	conduct 
	Transformer/self-attention  learning efficiently on the coarse graph level, such as graph patches~\cite{gao2022patchgt}, communities~\cite{kuangcoarformer} etc. 
	For example, in Coarformer~\cite{kuangcoarformer}, it first employs a graph coarsening technique to generate a global coarse graph view of the original graph and 
	then conducts Transformer on the coarse graph. 
	Similar strategy has also been employed in PatchGT~\cite{gao2022patchgt}. 
	Obviously, this strategy generally returns the coarse-level representation 
	which 
	fails to be fully aware of  original node representation in its learning process. 

	\section{Methodology}
	
\begin{figure*}[htpb]
		\centering
		\includegraphics[width=1.0\textwidth]{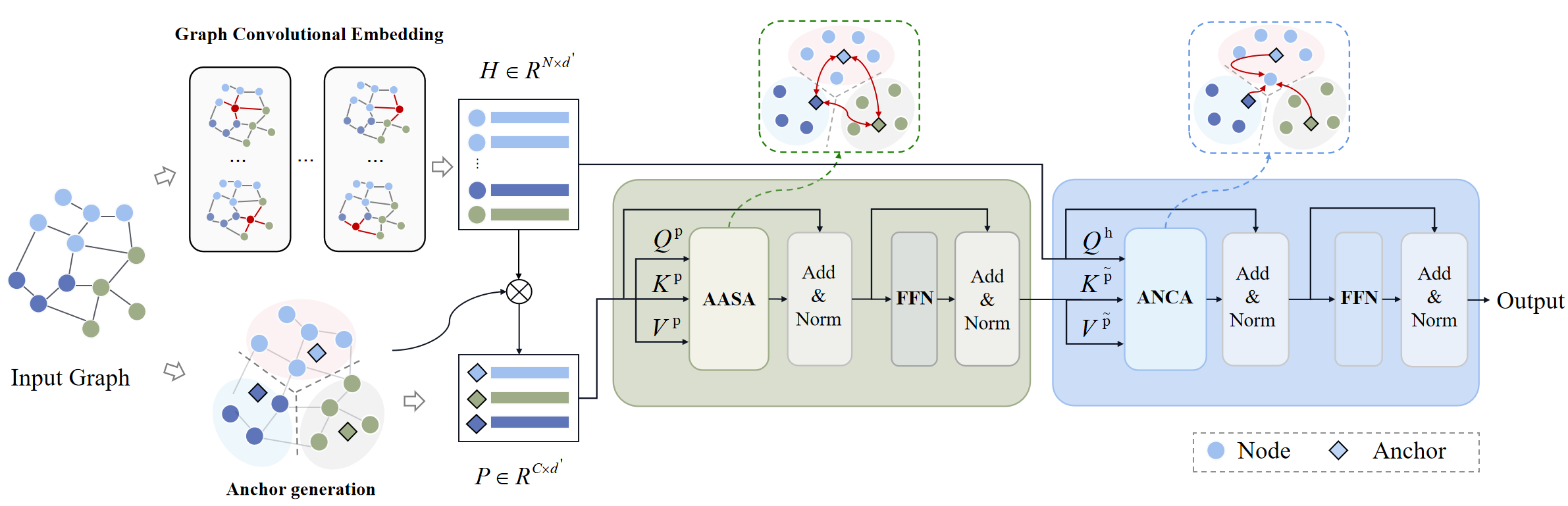}
		\caption{The architecture of Anchor Graph Transformer (AGFormer). 
			It mainly contains three modules: i) Graph Convolutional Emdedding module, ii) Anchor-to-Anchor Self-Attention (AASA) module and iii) Anchor-to-Node Cross-Attention (ANCA) module.
		}
		\label{fig:framework}
	\end{figure*}

	In this section, 
	we present our Anchor Graph Transformer (AGFormer) 
	for graph data representation learning. 
	As shown in Fig. \ref{fig:framework}, our AGFormer contains three main parts, i.e., 
	Graph Convolutional Embedding, 
	Anchor-to-Anchor Self-Attention (AASA) and 
	Anchor-to-Node Cross-Attention (ANCA). 
	We  introduce these modules in following subsections, respectively. 
	
	\subsection{Graph Convolutional Embedding}
	
	It is known that graph convolutional network 
	provides a fundamental module to 
	learn local neighbor-aware node embedding. 
	In our method, we adopt it to learn 
	the initial feature embeddings for graph nodes~\cite{NEURIPS2021_Graphormer,Wu2021GraphTrans,NguyenUGformer}. 
	We denote a given input graph as $\mathcal{G}(\mathcal{V},\mathcal{E})$, where $\mathcal{V} = \{v_1,\cdots,v_N\}$ is the set of $N$ nodes and $\mathcal{E}$ is the set of edges.
	The adjacency matrix $A\in \{{0,1}\}^{N\times N}$ represents  graph structure information, i.e., $A_{ij} = 1$ if node $v_i$ and $v_j$ are connected, otherwise $A_{ij} = 0$. We denote the node feature matrix as $X \in \mathbb{R}^{N\times d}$ where $d$ denotes the feature dimension. 
	To learn neighbor-aware node embeddings, we adopt multiple graph convolutional layers~\cite{GCN,GIN} on graph as 
	\begin{equation}
	Z = \mathrm{GCN}\bigl(X,A;\Theta\bigr)
	\end{equation}
	where $\Theta$ denotes the convolution parameters. 
	Here, many graph convolution network architectures can be adopted. 
	In this paper, we respectively use the commonly used graph convolution networks ~\cite{GCN,GIN}. 
	After obtaining $Z$, we perform a linear projection together with layer normalization (LN) to obtain the low-dimensional node embeddings  for the followed AGFormer module, i.e., 
	\begin{align}
	H = \mathrm{LN} \big( ZW^{Proj}\big )
	\end{align}
	where $W^{Proj} \in \mathbb{R}^{d \times d' }$ denotes the learnable projection matrix. 
	
	\subsection{Anchor Graph Transformer}
	
	The above graph convolution module generally fails to 
	capture the long-range dependencies of nodes. 
	Graph Transformers have been developed to address this issue. 
	The aim of Graph Transformers is to model the dependencies of all nodes via node-wise self-attention mechanism. 
	However, existing graph Transformers generally compute the `full' self-attention for all nodes, leading to high computational complexity. 
	In recent years, anchor-based graph model has been widely used in large-scale data mining problem, such as clustering~\cite{nie2021fast}, semi-supervised learning~\cite{wang2021fast} etc., to speed up the learning process. 
	Inspired by recent research on anchor graph techniques, 
	we develop a novel Anchor Graph Transformer (AGFormer) for graph representation. 
	The core idea of the proposed AGFormer is to  select a few representative anchors and convert node-to-node information propagation to anchor-to-anchor and anchor-to-node propagation, which thus makes it perform much more efficiently than regular Graph Transformers. 
	Overall, the proposed AGFormer mainly contains Anchor Generation, Anchor-to-Anchor Self-Attention (AASA) and Anchor-to-Node Cross-Attention (ANCA) steps, as introduced below. 

	\subsubsection{Anchor generation}
	
	The most straightforward way to select 
	anchors is to take cluster/community centers as anchors. 
	Many graph clustering algorithms can be
	adopted here. 
	In this paper, we use the commonly used Louvain~\cite{Louvain} algorithm which can adaptively obtain the communities for the input graph efficiently. 
	To be specific, using Louvain algorithm~\cite{Louvain}, we can adaptively obtain $C$ communities and the corresponding assignment matrix $S \in \mathbb{R}^{C \times N}$, where $S_{cj}=1$ denotes that the $j$-th node is assigned to the $c$-th community/cluster. 
	Then, we obtain an anchor node for each community by using the center representation of the community. 
	Let $P = \{p_1,p_2,\cdots,p_C\} \in \mathbb{R}^{C\times d'}$ denote the collection of
	feature representations of $C$ anchors. 
	Then, we can compute $P$ as follows:
	\begin{align}
	P = D^{-1}SH
	\end{align}
	where $D$ is the diagonal matrix with
	$D_{cc} = \sum_j S_{cj}$.
	$H$ is the initial features of graph nodes obtained via GCN, as shown in Eqs.(1,2). 
	

	\subsubsection{Anchor-to-Anchor Self-Attention}
	
	To achieve the information interaction among different anchors, we develop an Anchor-to-Anchor Self-Attention (AASA) module. 
	The core of this module is to capture the long-range dependencies of nodes through the information passing among different anchors.
	To be specific, as shown in Figure 2, 
	we first compute query $Q^p$, key $K^p$ and value $V^p$ by conducting three linear projections on $P$ respectively as 
	\begin{align}
	&Q^p = PW^p_1 \, \,\,\, K^p = PW^p_2, \,\,\, V^p = PW^p_3
	\end{align}
	where 
	$W^p_1, W^p_2$ and $W^p_3$ denote three linear projections. 
	Then, we apply self-attention on anchor nodes $P$, i.e., 
	\begin{align}
	&    {\mathrm{Attn}(Q^p, K^p, V^p)} = \mathrm{Softmax}\Bigl({\frac{Q^p {K^p}^T}{\sqrt{d^{'}}}}\Bigr) V^p
	\end{align}
	where 
	$d'$ indicates the dimension of the input node features.
	
	Finally, each anchor node updates its representation by aggregating  messages from other anchors and further conducting layer normalization and residual operation as 
	\begin{align}
	\label{pi}
	\Bar{P} = \mathrm{LN}\bigl( P + {\mathrm{Attn}(Q^p, K^p, V^p)}\bigr)
	\end{align}
	where $\mathrm{LN(\cdot)}$ refers to the layer normalization. 
	In addition, after obtaining anchor node representations, we utilize FFN (Feed Forward Network) which consists of two-layer MLP, to improve the expression ability of the network as follows
	\begin{align}
	\label{p}
	\widetilde{P} = \mathrm{LN}\bigl( \Bar{P} + {\mathrm{MLP}(\Bar{P},\Phi^{p})}\bigr)
	\end{align}
	where $\Phi^{p}$ denotes the learnable parameters of the FFN module.
	
	\subsubsection{Anchor-to-Node Cross-Attention} 
	
	After obtaining $\widetilde{P}$ (Eq.(\ref{p})) via anchor-to-anchor self-attention, we design the Anchor-to-Node Cross-Attention (ANCA) module to achieve message passing from anchors to each node. 
	
	To be specific, as shown in Figure 2, we first obtain query $Q^h\in\mathbb{R}^{N\times d'}$ by using linear projection on node features $H$ and compute key $K^{\tilde{p}}$ and value $V^{\tilde{p}}$ by conducting two linear projections on $\widetilde{P}$ respectively as 
	\begin{align}
	&Q^h = HW^h_1 \, \,\,\, K^{\tilde{p}} = \widetilde{P}W^{\tilde{p}}_2, \,\,\, V^{\tilde{p}} = \widetilde{P}W^{\tilde{p}}_3 
	\end{align}
	where 
	$W^{h}_1, W^{\tilde{p}}_2$ and $W^{\tilde{p}}_3$ denote three linear projections.  
	Then, we apply the cross-attention between anchors and nodes, i.e., 
	\begin{align}
	&    {\mathrm{Attn}(Q^h,K^{\tilde{p}}, V^{\tilde{p}})} = \mathrm{Softmax}\Bigl({\frac{Q^h {K^{\tilde{p}}}^T}{\sqrt{d^{'}}}}\Bigr) V^{\tilde{p}}
	\end{align}
	where 
	$\mathrm{Attn}(\cdot)$ denotes the attention function.
	Finally, 
	each node updates its representation by aggregating
	messages from all anchors and following layer normalization
	and residual operation as, 
	\begin{align}
	\label{h1}
	{\Bar{H}}= \mathrm{LN}\bigl(H + {\mathrm{Attn}(Q^h,K^{\tilde{p}}, V^{\tilde{p}})}\bigr)
	\end{align}
	The next step FFN (Feed Forward Network)  is further conducted on $\Bar{H}$ to obtain $\widetilde{H}$
	as 
	\begin{align}
	\label{h}
	\widetilde{H} = \mathrm{LN}\bigl( \Bar{H} + {\mathrm{MLP}(\Bar{H},\Phi^{h})}\bigr)
	\end{align}
	where $\Phi^{h}$ denotes the parameters of FFN.
	
	\subsection{Comparison with Related Works} 
	
	In this section, we compare our AGFormer with some other related graph Transformer methods which 
	include 
	GraphTrans\cite{Wu2021GraphTrans}, Coarformer~\cite{kuangcoarformer} and 
	PatchGT~\cite{gao2022patchgt}.  
	GraphTrans~\cite{Wu2021GraphTrans} adopts regular Transformer architecture for graph representation in which node-to-node self-attention is employed to capture the long-range dependences of nodes. 
	In contrast, AGFormer  converts node-to-node message passing into an anchor-to-anchor and anchor-to-node message
	passing process which is obviously more efficient than GraphTrans~\cite{Wu2021GraphTrans}, as further validated in Experimental section.  
	In Coarformer~\cite{kuangcoarformer}, it first employs a graph coarsening technique to generate the global coarse graph view of the original graph and 
	then conducts Transformer on the coarse graph. 
	Similar strategy has also been employed in PatchGT~\cite{gao2022patchgt}. 
	Obviously, these methods generally return coarse-level representations 
	that 
	fail to be fully aware of  original node representations in their learning process. 
	Differently, 
	our AGFormer 
	involves both anchor-to-anchor and 
	anchor-to-node message passing modules which can learn 
	discriminative node-level representations for graph data.

	\section{Experiment}
	In this section, we empirically investigate the effectiveness and advantages of AGFormer on several graph classification benchmark datasets and compare our method with some other related works. 
	\subsection{Experiment setup}
	\textbf{Datasets.}  
	First, we conduct experiments on three bioinformatics datasets including NCI1~\cite{wale2008comparison}, NCI109~\cite{wale2008comparison} and MUTAG~\cite{kriege2012subgraph}.
	In these datasets, each graph represents a compound in chemical molecules whose  nodes represent atoms and edges denote bonds. 
	We also evaluate AGFormer on two social network datasets including COLLAB and IMDB-BINARY (IMDB-B)~\cite{yanardag2015deep}. 
	COLLAB is derived from three public collaborative datasets (High Energy Physics, Condensed Matter Physics, and Astrophysics)~\cite{leskovec2005graphs}. It is a scientific collaborative dataset, representing collaborative  relationships among authors.
	Each node represents a researcher and edges denote the collaborations between researchers. 
	IMDB-BINARY is the movie collaboration dataset. Each graph is derived from a predesignated movie genre, where each node represents an actor and each edge represents whether two actors appearing in the same movie. 
	The statistics of these datasets are summarized in Table~\ref{tab:datasets}.
	\begin{table}[htbp]
		\centering
		\caption{The statistics of all datasets.}
		\label{tab:datasets}
		\begin{tabular}{cccccc}
			\toprule
			\textbf{Datasets} & \textbf{NCI1} & \textbf{NCI109} & \textbf{MUTAG} & \textbf{COLLAB} & \textbf{IMDB-B}\\
			\midrule
			Graphs & 4110 & 4127 & 188 & 5000 & 1000\\
			Avg. Nodes & 29.87 & 29.68 & 17.93 & 74.49 & 19.77\\
			Avg. Edges & 32.30 & 32.13 & 19.79 & 2457.78 & 96.53\\
			Max. Nodes & 111 & 111 & 28 & 492 & 136\\
			\midrule
			Classes & 2 & 2 & 2 & 3 & 2\\
			\bottomrule
		\end{tabular}
	\end{table}
	
	\textbf{Comparison Methods.} 
	To demonstrate the effectiveness of AGFormer,
	we first compare it with three graph kernel methods, including  Weisfeiler-Lehman subtree kernel (WL subtree)~\cite{WLSK},  Random
	Walk Graph Kernel (RWGK)~\cite{RWGK} and  Shortest Path kernel based on Core variants (CORE SP)~\cite{nikolentzos2018degeneracyCORESP}. 
	Then, we compare our method with five graph representation methods, including  Graph Isomorphism Network (GIN)~\cite{GIN},  High-Order Graph Convolution Network (HO-GCN)~\cite{HOGCN},   Dual Attention Graph Convolution Network (DAGNN)~\cite{Chen2019DAGCN},  Graph Multiset Transformer (GMTPool)~\cite{GMTPool} and  Graph Capsule Network
	(GCAPS-CNN)~\cite{verma2018graphGCCNN}. 
	Finally, we compare our AGFormer with some current graph Transformers, i.e., Universal Graph Transformer (UGformer)~\cite{nguyen2022UGformer},
	two variants of  Graph Transformer (GraphTrans)~\cite{Wu2021GraphTrans}, i.e., GraphTrans (GCN) and GraphTrans (GIN).
	We also provide the results of vanilla Transformer~\cite{vaswani2017Transformer}  in experiments.
	For most of methods~\cite{WLSK,RWGK,GIN,Chen2019DAGCN,verma2018graphGCCNN,zhang2020improvingGAT-GC,nguyen2022UGformer}, all results are referenced from their own published papers.
	For GraphTrans~\cite{Wu2021GraphTrans} and Transformer methods~\cite{vaswani2017Transformer}, we directly report the results provided in previous work~\cite{Wu2021GraphTrans} on NCI1 and NCI109 datasets and obtain the results on other datasets by running their provided codes with the same experimental setting as our method.
	
	\begin{table*}[!htp]
		\begin{center}
			\caption{Comparison of different methods on five datasets. The best, second and third results are marked by red, blue and green respectively.   ${}^\dagger$ indicates the results we reproduced.}
			\label{tab:summary}
			\begin{tabular}{c|c|ccccc}
				\toprule
				\hline
				\multicolumn{2}{c|}{\textbf{Methods}}& \textbf{NCI1(\%)} &\textbf{NCI109(\%)} &\textbf{MUTAG(\%)} & \textbf{COLLAB(\%)} & \textbf{IMDB-BINARY(\%)}\\
				\hline
				\multirow{3}{*}{\textbf{Kernel}} & WLSK~\cite{WLSK} & $82.19\pm0.18$ & \color{blue}$82.46\pm0.24$ & $82.05\pm0.36$ & $77.39\pm0.35$ & $71.88\pm0.77$\\
				&CORE SP~\cite{nikolentzos2018degeneracyCORESP}&$73.46\pm0.32$&-&$88.29\pm1.55$ &-& $72.62\pm0.59$\\
				&RWGK~\cite{RWGK}&-&-&$80.77\pm0.72$ &-& $67.94\pm0.77$\\
				\hline
				\multirow{4}{*}{\textbf{Graph Representation}}&GIN~\cite{GIN}&\color{green}$82.70\pm1.70$&-&\color{green}$89.40\pm5.60$ &$80.20 \pm 1.90$ & \color{blue}$75.10\pm 5.10$\\
				&DAGCN~\cite{Chen2019DAGCN}&$81.68\pm1.69$&$81.46\pm1.51$&$87.22\pm6.10$ &-&-\\
				&GMTPool~\cite{GMTPool} & - & - & $83.44\pm1.33$ & \color{green}$80.74\pm0.54$ & $73.48\pm0.76$\\
				&GCAPS-CNN~\cite{verma2018graphGCCNN}& \color{blue}$82.72\pm2.38$ & $81.12\pm1.28$ & - & $77.71\pm2.51$ & $71.69\pm3.40$\\
				\hline
				\textbf{Transformer} &Transformer~\cite{vaswani2017Transformer}& $68.50\pm2.60$ & $70.10\pm2.30$ & ${83.75\pm8.50}^\dagger$ & ${76.50\pm0.84}^\dagger$ & ${71.20\pm3.66}^\dagger$\\
				\hline
				\multirow{4}{*}{\textbf{GNN+Transformer}}
				&UGformer~\cite{nguyen2022UGformer}&-&-&\color{blue} $89.97\pm3.65$ & $77.84\pm1.48$ & \color{red}$77.05\pm3.45$\\
				&GraphTrans(GCN)~\cite{Wu2021GraphTrans} & $81.30\pm1.90$ & $79.20\pm2.20$ & $87.22\pm7.05$ & ${81.59\pm1.48}^\dagger$ & ${74.10\pm3.11}^\dagger$\\
				&GraphTrans(GIN)~\cite{Wu2021GraphTrans} & $82.60\pm1.20$ & $82.30\pm2.60$ & $89.24\pm5.29$ & ${81.68\pm1.73}^\dagger$ & ${74.50\pm2.89}^\dagger$\\
				\cline{2-7}
				&\textbf{AGFormer(GCN)}& $ 82.38\pm2.13$ & $\color{green}{82.33}\pm{1.41}$ & \color{red}$90.00\pm5.44$ & \color{red}$82.42\pm1.32$ & \color{green}$74.90\pm4.18$\\
				&\textbf{AGFormer(GIN)}& $\color{red}{83.58}\pm1.81$ & $\color{red}{83.40}\pm{1.23}$ & $88.78\pm8.78$ & \color{blue}$81.88\pm0.98$ & $74.00\pm4.52$\\
				\toprule 
			\end{tabular}
		\end{center}
	\end{table*}
	
	\textbf{Implementation Details.} 
	The proposed AGFormer consists of two main parts, i.e., multi-layer GCN and the proposed Transformer.
	In our experiments, we use GCN~\cite{GCN}, GIN~\cite{GIN} as our GNN backbone to extract neighbor-aware representations respectively, namely AGFormer (GCN) and AGFormer (GIN). 
	For dataset NCI1~\cite{wale2008comparison}, MUTAG~\cite{kriege2012subgraph}, we use four-layer GNN. For dataset NCI109~\cite{wale2008comparison}, COLLAB~\cite{leskovec2005graphs} and IMDB-B~\cite{yanardag2015deep}, we use five-layer GNN. 
	We set the number of  hidden units in GNN to $256$. 
	For information interaction module, it consists of one layer of anchor self-attention module and one layer of anchor-to-node self-attention module. 
	We set the number of  hidden units in the proposed AGFormer module to $256$. 
	The dropout rate is set to $0.1$ on most datasets and set to $0.2$ on COLLAB dataset~\cite{leskovec2005graphs} for AGFormer (GIN).
	Similar to previous work~\cite{Wu2021GraphTrans}, 
	we jointly optimize graph convolution module and AGFormer together by minimizing the cross-entropy loss with the Adam optimizer~\cite{Adam}.
	We set the learning rate and weight decay of both two parts to $0.0001$. 
	On all datasets, we train our AGFormers with $100$ epochs.
	The batch size on dataset NCI1~\cite{wale2008comparison} and NCI109~\cite{wale2008comparison} is set to $256$. For dataset MUTAG~\cite{kriege2012subgraph}, COLLAB~\cite{leskovec2005graphs} and IMDB-B~\cite{yanardag2015deep}, the batch size is set to $128$. 
	We evaluate our model by using 10-fold cross-validation and report the average accuracy with standard deviation on the testing set. 
	In our method, we use the 
	Louvain algorithm~\cite{Louvain}
	to obtain anchors. The number of anchors $C$ is determined adaptively in Louvain algorithm~\cite{Louvain}. 
	
	\subsection{Comparison Results} 
	Table~\ref{tab:summary} shows the experimental results of our proposed AGFormer model on five datasets, i.e., NCI1, NCI109, MUTAG, COLLAB and IMDB-B. 
	Here, we can observe that 
	1) Our AGFormer performs better than some other recent Graph Transformers including standard Transformer~\cite{vaswani2017Transformer}, GraphTrans~\cite{Wu2021GraphTrans} and UGFormer~\cite{nguyen2022UGformer}. Compared with baseline method GraphTrans~\cite{Wu2021GraphTrans}, the accuracy of our method is improved by about 1.0\% on five datasets on average. This clearly demonstrates that the proposed AGFormer is more effective on graph data representation by taking advantage of high-level anchor node information. 
	2) Our proposed AGFormer consistently outperforms some other graph representations on five datasets. For example, the average improvement is 3.22\% compared to the GMTPool~\cite{GMTPool} model. This further demonstrates the effectiveness of the proposed method by  capturing the long-range dependencies of nodes on graph learning tasks. 
	3) Comparing with traditional graph kernel methods, our proposed method performs obviously better on most datasets 
	which further demonstrates the effectiveness of the proposed graph Transformer model on addressing graph data learning tasks. 

	\begin{figure}[htpb]
		\centering
		\subfigure[Using GCN as backbone]{\includegraphics[width=1.7in]{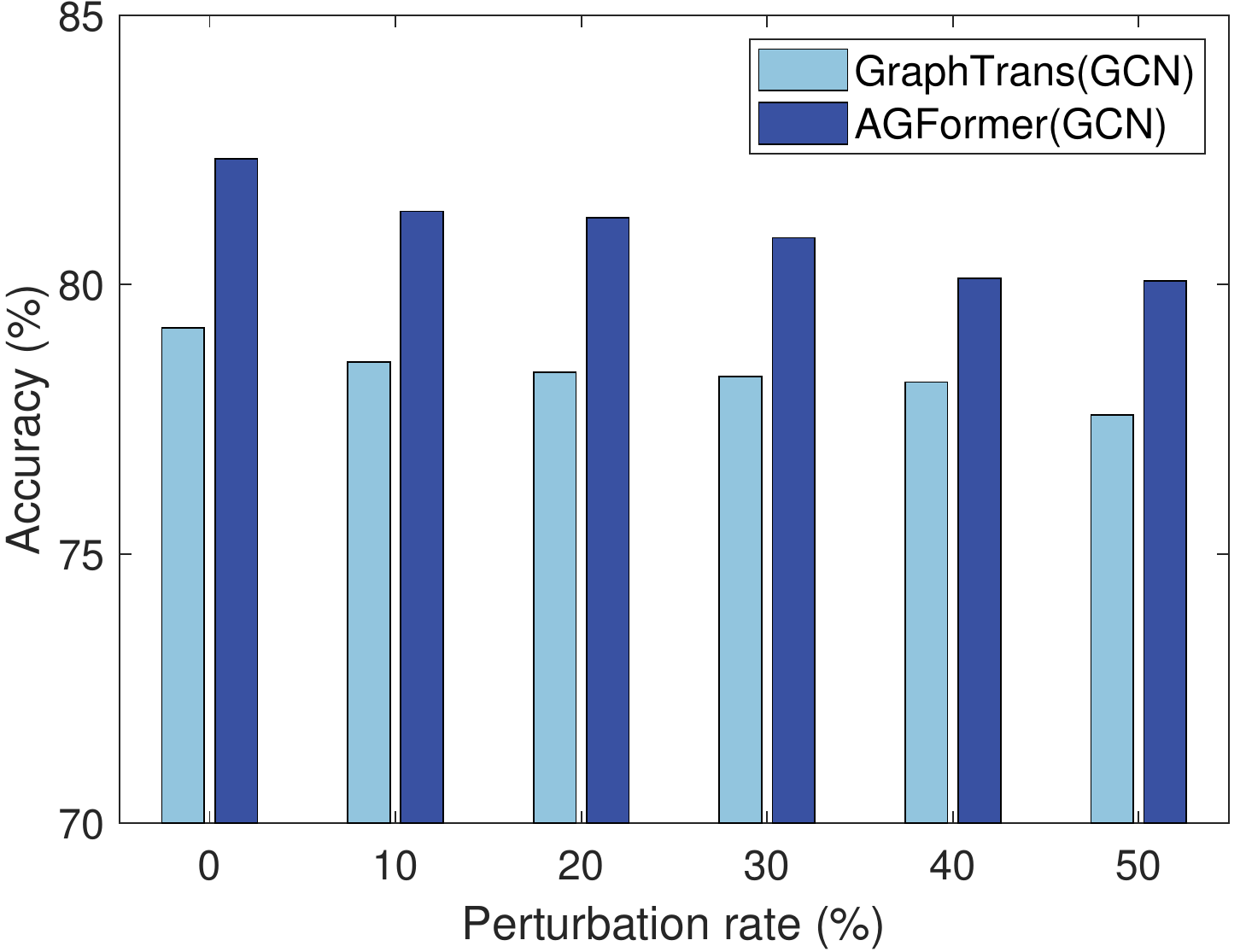}}
		\subfigure[Using GIN as backbone]{\includegraphics[width=1.7in]{flipgcn-eps-converted-to.pdf}}
		\caption{Robustness performance of AGFormer on dataset NCI109 under different perturbation rates.}
		\label{fig:Robustness}
	\end{figure}
	
	\subsection{Model analysis}
	\subsubsection{Robustness analysis}
	We investigate the robustness of AGFormer by generating perturbed graphs using the global attack method, i.e., random attack perturbs the graph structure by randomly flipping fake edges with different probabilities. 
	The experimental accuracies under different disturbance probabilities are shown in  Figure \ref{fig:Robustness}. 
	It can be observed that AGFormer consistently outperforms the baseline GraphTrans~\cite{Wu2021GraphTrans} on the attacked graph data. 
	This clearly demonstrates  that our AGFormer performs obviously more robustly than baseline method GraphTrans~\cite{Wu2021GraphTrans} w.r.t graph attacked noises. 

	
	
	\begin{figure}[!htbp]
		\centering
		\subfigure{\includegraphics[width=2.5in]{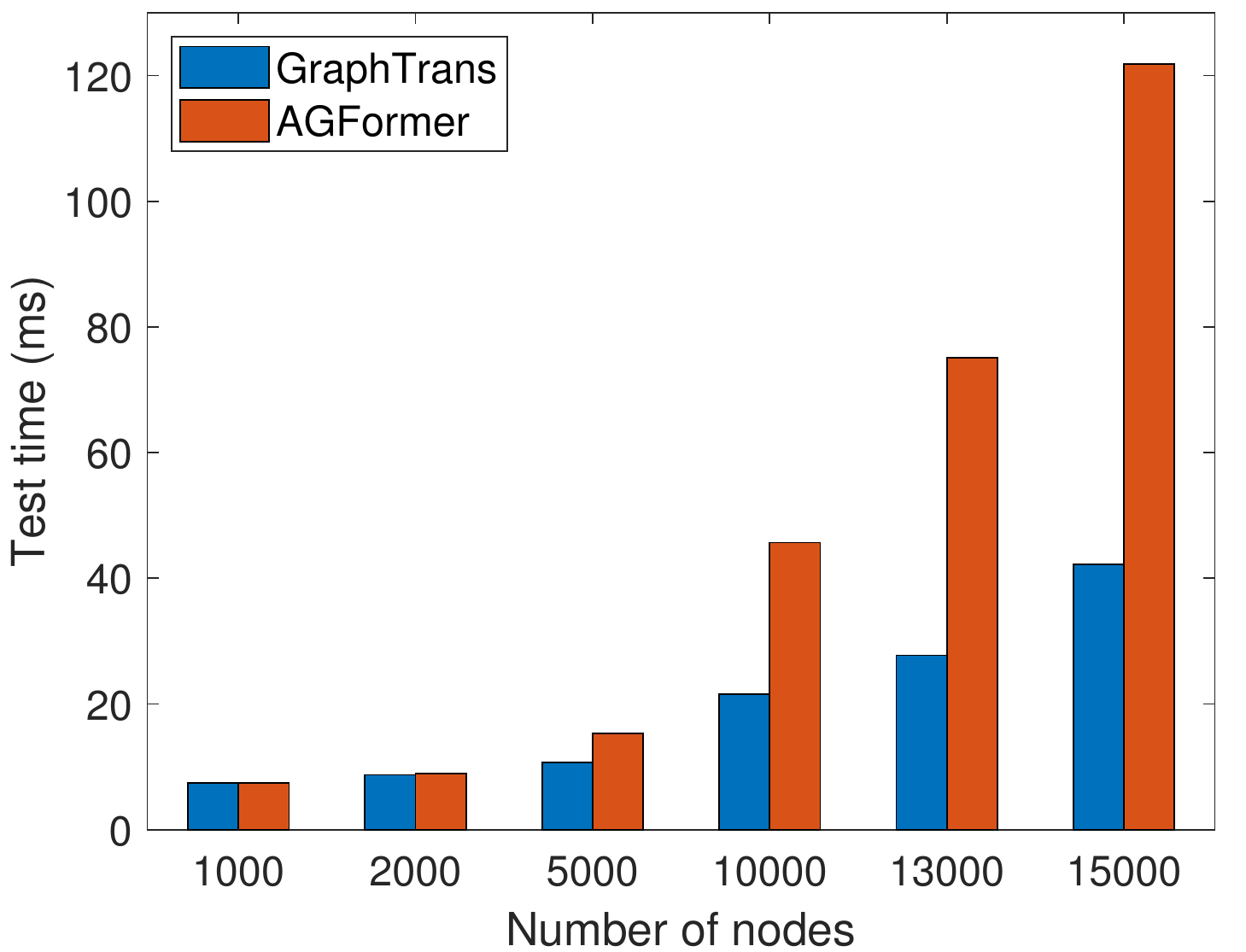}}
		\caption{Time (ms) comparison of AGFormer and GraphTrans on different simulation graph data.}
		\label{fig:Efficiency}
	\end{figure}
	
	\subsubsection{Efficiency analysis}
	To verify the efficiency of our AGFormer model, we randomly generate different sizes of simulated graphs with edge rate $1\%$  
	and calculate the test running time of  AGFormer and GraphTrans~\cite{Wu2021GraphTrans} on these simulated data. 
	For fair comparison, both 
	AGFormer and GraphTrans~\cite{Wu2021GraphTrans} adopt the same network settings. 
	Figure~\ref{fig:Efficiency} demonstrates the testing time of AGFormer and baseline GraphTrans~\cite{Wu2021GraphTrans} across different sizes of simulated graph data.
	Here, we can observe that 
	as the graph size increases, our AGFormer performs obviously more efficiently than baseline method GraphTrans~\cite{Wu2021GraphTrans}. This 
	clearly demonstrates the efficiency of the proposed AGFormer (especially on large-scale graph)  by 
	leveraging anchor graph model into graph Transformer designing. 
	

	
	\subsubsection{Parameters analysis}
	
	In this paper, we select anchors by using
	cluster/community centers via  Louvain algorithm~\cite{Louvain}. 
	The  benefit 
	of this algorithm is to 
	generate anchors 
	automatically.  
	Empirically, the number of 
	anchors generated 
	by Louvain algorithm is generally about $30\%$-$50\%$ of graph size on all used datasets in experiments.   To evaluate the effectiveness of this strategy, we further test our method with random anchor selection.  
	Table\ref{tab:anchorSelection} 
	shows the comparison results of these two anchor generation methods. 
	We can observe that 
	AGFormer with random anchors can also 
	return feasible solution. 
	The cluster center based anchor strategy is obviously beneficial for AGFormer.

	\begin{table}[htbp]
		\centering
		\caption{Comparison results of two anchor selection methods (random vs. Louvain).}
		\begin{tabular}{c|c|cc}
			\toprule
			\hline
			\textbf{Method} & - & \textbf{NCI1} & \textbf{NCI109}  \\
			\hline
			\multirow{2}{*}{\textbf{AGFormer(GCN)}} & Random & $81.05\pm1.50$ & $80.78\pm1.81$\\
			
			& Louvain & $82.38\pm2.13$ & $82.33\pm1.41$\\
			\hline
			\multirow{2}{*}{\textbf{AGFormer(GIN)}} & Random & $79.49\pm1.38$ & $81.53\pm1.60$\\
			& Louvain & $83.58\pm1.81$ & $83.40\pm1.23$\\
			\toprule
		\end{tabular}
		\label{tab:anchorSelection}
	\end{table}

	\section{Conclusion}
	
	This paper proposes a novel Anchor Graph Transformer
	(AGFormer) for efficient and robust graph data represe learning. AGFormer first obtains some representative
	anchors and then converts node-to-node message passing into
	anchor-to-anchor and anchor-to-node message passing process. 
	AGFormer provides an efficient and robust way to
	learn node-level representations by integrating local and global
	dependences together. 
	Extensive experiments on several widely used datasets demonstrate the effectiveness and benefits (efficiency, robustness) of proposed AGFormer.

	\bibliography{reference_list}
	\bibliographystyle{ieeetr}

	\ifCLASSOPTIONcaptionsoff
	\newpage
	\fi
	
\end{document}